\pdfoutput=1
\documentclass[11pt]{article}
\usepackage[]{acl}
\usepackage{times}
\usepackage{latexsym}
\usepackage{amsmath}
\usepackage{graphicx}
\usepackage{hyperref}
\usepackage[T1]{fontenc}
\usepackage[utf8]{inputenc}
\usepackage{microtype}

\title{The COVID That Wasn't: Counterfactual Journalism Using GPT}

\author{Sil Hamilton \\
  McGill University\\
  \texttt{sil.hamilton@mcgill.ca} \\\And
  Andrew Piper \\
  McGill University\\
  \texttt{andrew.piper@mcgill.ca} \\}

\begin{document}
\maketitle
\begin{abstract}
In this paper, we explore the use of large language models to assess human interpretations of real world events. To do so, we use a language model trained prior to 2020 to artificially generate news articles concerning COVID-19 given the headlines of actual articles written during the pandemic. We then compare stylistic qualities of our artificially generated corpus with a news corpus, in this case 5,082 articles produced by CBC News between January 23 and May 5, 2020. We find our artificially generated articles exhibits a considerably more negative attitude towards COVID and a significantly lower reliance on geopolitical framing. Our methods and results hold importance for researchers seeking to simulate large scale cultural processes via recent breakthroughs in text generation.
\end{abstract}

\section{Introduction}
The rush to cover new COVID-19 developments as the virus spread across the world over the first half of 2020 induced a variety of editorial mandates from public broadcasters. Chief among these was a desire to mitigate shock from a public unaccustomed to large-scale public health emergencies of the calibre COVID-19 presented. This desire translated into systematic underreporting together with a reluctance to portray COVID-19 as the danger it was \citep{quandt2021stooges, boberg_pandemic_2020}. Broadcasters in the United States \citep{zhao_media_2020}, the United Kingdom \citep{garland2021consensus}, and Italy \citep{solomon_journalistic_2021} all exhibited this phenomenon.

Although many studies have verified the above effects, few if any studies to date have considered \emph{alternative} approaches the media could have taken in their portrayal of COVID-19. Evaluating these alternatives is critical given the close relationship between media framing, public opinion, and government policy \citep{ogbodo_communicating_2020, lopes_paranoia_2020}.

In this paper we present a novel method of simulating media coverage of real world events using Large Language Models (LLMs) as a means of interpreting news industry biases. LLMs have been used in a variety of settings to generate text for real-world applications \citep{meng22_gener_train_data_with_languag_model, drori21_neural_networ_solves_gener_mathem}. To our knowledge, they have not yet been used as a tool for critically understanding the interpretation of events through media coverage or other forms of cultural framing.

To do so, we use Generative Pre-trained Transformer 2 (GPT-2), which was trained on text produced prior to the onset of COVID-19, to explore how the Canadian Broadcasting Corporation (CBC) covered COVID and how else they might have reported on these breaking events. By generating thousands of simulated articles, we show how such ``counterfactual journalism'' can be used as a tool for evaluating real-world texts.

\section{Background}
The COVID-19 pandemic has given researchers a variety of opportunities to study human behavior in response to a major public health crisis. One core dimension of this experience is reflected in the changing role that the media has played in communicating information to the public in a quickly changing health environment \citep{van2021covid, lilleker2021political}. Times of crises enshrine the media as a valuable mediator between the public and government.

This changing role registers itself in the editorial policies at news corporations across the world. Research published over the past two years has confirmed that public news broadcasters in Australia, Sweden, and the United Kingdom all significantly altered their editorial style in response to both societal and governmental pressures \citep{lewis_mapping_2021, shehata2021swedish, birks2021just} during COVID-19.

What this research has so far lacked is the ability to infer what \emph{could} have been communicated, i.e. what losses were entailed in these editorial shifts. While a great deal of recent work has studied the biases intrinsic to large language models\footnote{See \citet{garrido-munoz_survey_2021} for a recent survey of works investigating latent biases present within large language models.}, no work to date has used LLMs to study the biases of human generated text. Reporting on real-world events inevitably requires complex choices of selection and evaluation, i.e. which events and which actors to focus on along with modes of valuation surrounding those choices. Simulating textual production given similar prompts such as headlines can provide a means of better understanding the editorial choices made by news agencies.

In using a language model as a simulative mechanism, we draw on a long research tradition of using simulation to understand real-world processes. Simulation has proven a boon for those working in the sciences, including climate science and physics \citep{winsberg2010science}, and for those working in the social sciences, where agent-based social modelling has led to advances in understanding complex social phenomena \citep{squazzoni2014social}. We seek to bring these techniques to the study of cultural behavior, where simulation has historically seen less of an uptake \citep{manovich_science_2016}.

\section{Method}
Our project consists of the following principal steps:

\begin{enumerate}
\item Create a news corpus drawn from our target time-frame (15 January to 5 May 2020) whose content is COVID-19 related.
\item Fine-tune a language model whose generative output is statistically similar to a random sampling of our news source published \emph{before} our target time-frame, i.e. prior to COVID.
\item Using this model, generate full-length text articles using various prompts, including headlines and associated metadata.
\item Compare generated text articles with the original news corpus across key stylistic metrics.\footnote{We make our code available  \href{https://git.sr.ht/~srhm/counterfactual-generation-scripts}{here.}}
\end{enumerate}

\subsection{Corpus}
We first obtain a comprehensive collection of CBC News' online articles concerning COVID-19 published between January and May 2020 from Kaggle \citep{han_covid-19_nodate}.  Our corpus contains 5,114 articles all in the form of a headline, subheadline, byline, date published, URL, and article text. Deduplicating and cleaning the corpus with a series of \verb.regex. filters leaves 5,082 articles spread across the first four months of COVID-19.

\subsection{Language Model}
We use a Transformer-based large language model (LLM) as our CBC simulacrum. We formalize our model as follows: we define an article as a chain of $k$ tokens. Let $X(d, \theta)$ be a probability distribution representing the pulling of a token out from the language model, where $d$ is the article metadata and $\theta$ are the prior weights. The probability of drawing $k$ tokens is then
\begin{equation}
  \Pr(\bar{x}_k) = \prod_{i=1}^k \Pr(X(d, \theta) = x^i | \bar{x}_i)
\end{equation}
where $\bar{x}^i$ is the $i^{th}$ element of the vector $\bar{x}$, and $\bar{x}_i$ is the vector consisting of the first $i$ elements of the vector $\bar{x}$.

Selecting a pretrained language model suitable for use as a base with which to further train with specific writing samples is a non-trivial task given the plurality of large language models released in the past four years \citep{hugging_face_models_nodate}. We surveyed models for candidates possessing the following qualities:

\begin{itemize}

\item the model must be neither egregious nor lacking in parameter count;
\item domain-relevant samples must have been present in the pretraining corpus;
\item and most importantly, the model must not be aware of COVID.
\end{itemize}

Keeping with the above requirements, we select the medium-sized Generative Pre-trained Transformer-2 (GPT-2) as distributed by OpenAI as our candidate model. We found the medium-sized GPT-2 model desirable because it is light enough to be fine-tuned with a single consumer-level GPU; CBC News was the 21\textsuperscript{st} most frequent data source OpenAI used in producing its training set \citep{clark_gpt-2_2022} and the model was trained in 2018, two years before the beginning of COVID-19.

\subsubsection{Fine-tuning}
Provided with sufficient context in the prompt, a freshly obtained GPT-2 model produces qualitatively convincing news article text. It will, however, periodically confuse itself with exactly which publication it is imitating, e.g. it can switch from sounding like CNN to CBC to BBC in a single text. For the purposes of comparison with a single news source, it is thus necessary to fine-tune the model with example texts encapsulating the desired editorial and writing style.

Fine-tuning is a two-step process. We first gather a sequence of texts best representing our target writing mode before fine-tuning a stock GPT-2 model with the training dataset.

\paragraph{Training Dataset}
We use a web scraper to extract a random selection of news articles published between 2007 and 2020 from CBC News' website. We configure our scraper to pull the same metadata as our COVID-19 dataset: headline, subheadline, date, URL, and article text. We again deduplicate to reduce the possibility of overfitting our model. With this method we collect 1,368 articles with an average length of 660 words per article.

We next construct a dictionary structure to formalize both our generation targets and to provide GPT-2 a consistent interface with which to aid it in logically linking together pieces of metadata. Previous research has indicated fine-tuning LLMs with structured data aids the model in both understanding and reacting to meaningful keywords \citep{ueda_structured_2021}. We therefore structure our fine-tuning data in a dictionary. We provide a template of our structure below.
\begin{verbatim}
{
 'title': 'Lorem ipsum…',
 'description': 'Lorem ipsum…',
 'text': 'Lorem ipsum…'
}
\end{verbatim}
We produce one dictionary per article in our training set. We convert each dictionary to a string before appending it to a final dataset text file with which we train GPT-2.

\paragraph{Training}
With our training dataset in hand, we proceed to configure our training environment. We use an Adam optimizer with a learning rate of $2e^{-4}$ and run the process \citep{https://doi.org/10.48550/arxiv.1412.6980}. Training the model for 20,000 steps over six hours results in a final model achieving an average training loss of 0.10.

\subsubsection{Model Hyperparameters}
\label{sec:hyperparameters}
In addition to fine-tuning our model, we experiment with different hyperparameters and prompt strategies. Numerous prior studies have described the effects hyperparameter tuning has on the token generation process \citep{van_stegeren_fine-tuning_2021, xu_systematic_2022}. For our purposes, we use three prompting strategies when generating our synthetic news articles along with one further parameter (\emph{temperature}):

\paragraph{Standard Context}
Only title and description metadata are used as context $d$ for the model.

\paragraph{Static Context}
In addition to the standard context, we supply the model with an additional \verb.framework. key containing a brief description of the COVID-19 pandemic found on the website of the Centre for Disease Control (CDC) in May 2020. All generation iterations use the same description.

\paragraph{Rolling Context}
We again supply the model with an additional \verb.framework. key, but keep the description of COVID-19 contemporaneous with the date of the real article in question. We again use the CDC as a source but instead use the Internet Archive's Way Back Machine API to scrape dated descriptions.\footnote{https://archive.org/help/wayback\_api.php}

\paragraph{Temperature} We manipulate the temperature hyperparameter during generation with half-percentage steps shifting the temperature between $0.1…1$. The temperature value is a divisor applied on the \verb.softmax. operation on the returned probability distribution, the affect of which effectively controls the overall likelihood of the most probable words. A high temperature results in a more dynamic and random word choice, while a lower temperature encourages those words which are most likely according to the model's priors.

\paragraph{Models} Manipulating the above hyperparemeters gives us the following model framework:

\begin{itemize}
    \item Model 1: headline-only, temperature between 0.1 and 1
    \item Model 2: static context, temperature between 0.1 and 1
    \item Model 3: rolling context, temperature between 0.1 and 1
\end{itemize}

We find that manipulating the \verb.softmax. temperature hyperparameter has no measurable effect on our measures described below. We thus proceed using only three primary models for article generation using a temperature of $0.50$, which we refer to in the remainder of the paper as Models 1, 2, and 3. When we rely on a single model to exhibit results Model 3 will be the model we choose to illustrate.

\subsection{News Article Generation}
\label{sec:generation}
Having now obtained both our models and our real-world corpus, we proceed with the text-generation step by prompting our model with metadata taken from CBC's COVID-19 articles. The generation process takes the following form:

\begin{enumerate}
\item For each article dictionary, extract the \verb.title. and \verb.description. keys.
\item For each pair of keys, create a new dictionary and add an empty \verb.text. key.
\item Convert all new dictionaries to strings and tokenize using GPT-2's Byte Pair Encoder (BPE).
\item Have GPT-2 predict the contents of the \verb.text. key using the title and description as context, generating 750 tokens in the process.
\item Collect generated tokens and insert into the \verb.text. key.
\end{enumerate}

Doing so generates 5,082 counterfactual news articles that temporally correspond to our CBC News corpus for each of our three primary models.  Each article pair in each model thus shares metadata but differs in content, with one being original and the other generated.

We provide here examples drawn from our simulated-actual article pairs to illustrate the performance of our models. Further examples may be found in \autoref{sec:appendix}.

\paragraph{Headline 1:} "China confirms human-to-human transmission of new coronavirus."

\subparagraph{CBC:} "Human-to-human transmission has been confirmed in an outbreak of a new coronavirus, the head of a Chinese government expert team said Monday, as the total number of cases more than tripled and the virus spread to other cities in China."

\subparagraph{GPT:} "An outbreak of a new coronavirus has been confirmed in southern China's Hebei Province, the lead author of a scientific paper said Thursday. The total number of cases more than triples the number of cases in the area, which corresponds to the Beijing and Shanghai hot spots..."

\paragraph{Headline 2:} "Quebec travel agencies feel the heat as local travellers cancel flights to China."

\subparagraph{CBC:} "[Name omitted for privacy] has dreamed of heading to Thailand with her partner for years. But with a five-month-old baby in tow and 14 cases of the coronavirus reported in the area so far..."

\subparagraph{GPT:} "Quebec travel agencies are feeling the heat as local travellers cancel flights to China. China remains the most dangerous place on Earth for travellers..."

\subsection{Measures}
To assess the stylistic differences between our simulated and real-world corpora we use the following measures:

\paragraph{Measure 1: Sentiment}
We measure the sentiment of each article with the open-source Python library \textit{VADER}\@ \citep{hutto2014vader}. Prior studies have validated the use of VADER on journalistic texts, finding the model to be superior to various alternatives in detecting sentiment \citep{10.1007/978-3-030-77517-9_9}. We additionally validate a small sample of measured sentences to ensure the accuracy of the tool.

We measure sentiment by first splitting a given article into $s$ sentences, obtain the compound polarity score ($-1…1$) for each $s$, then average all $s$ into a final score for the article. The resulting real number represents the overall sentiment of the article.

We apply a number of heuristics to ensure the sentiment score accurately reflects the reality of COVID-19. Words that would previously represent a positive sentiment (such as a “‘positive’ test”) become negative in actuality during a pandemic. It is the same for certain negative terms like “testing ‘negative.’” Our heuristics appropriately shift such terms as they appear, allowing for a more accurate measurement. As we later show, our heuristics demonstrate a strong correlation between our simulated and actual corpora.

\paragraph{Measure 2: Named Entity Recognition}
We detect and track named entities in each article with the use of the Python library \textit{spaCy} and their \verb.en_core_web_sm. model \citep{montani_explosionspacy_2022} given prior studies found the model is effective in recognizing named entities \citep{8931850}. We specifically tally all entities tagged as being a person, geopolitical entity, or organization on an intra-article basis.
\paragraph{Measure 3: Focus}
We take the ratio of total unique named entities $e$ over article length $l$ and call it \textit{focus}, a novel measure for how focused a given article $x$ is around a given set of entities:
\begin{equation}
focus(x) = \frac{e}{l}
\end{equation}
We see focus as a measure of concentration around prominent agents in the news.

{\renewcommand{\arraystretch}{1}
\begin{table}
\centering
\begin{tabular}{| p{0.7\linewidth} | p{0.2\linewidth} |}
\hline
\textbf{Sentence} & \textbf{Sentiment} \\
\hline
“Thousands of cyclists pedalled along empty Toronto highways today, enjoying the good weather and raising money for charity.” & 0.8074 \\
“'They're good at running them and we have to create the right environment for them,' she said.” & 0.6124 \\
“She said it's not good enough to say there's a strategy — that the province needs a strategy in action.” & -0.3412 \\
“Transportation Minister Clare Trevena said the incident is 'obviously' worrisome.” & -0.4019 \\
\hline
\end{tabular}
\caption{A subset of sentences and their VADER sentiment score from the control dataset.}
\label{table:sentences}
\end{table}}
\begin{figure}
  \centering
  \includegraphics[width=1\linewidth]{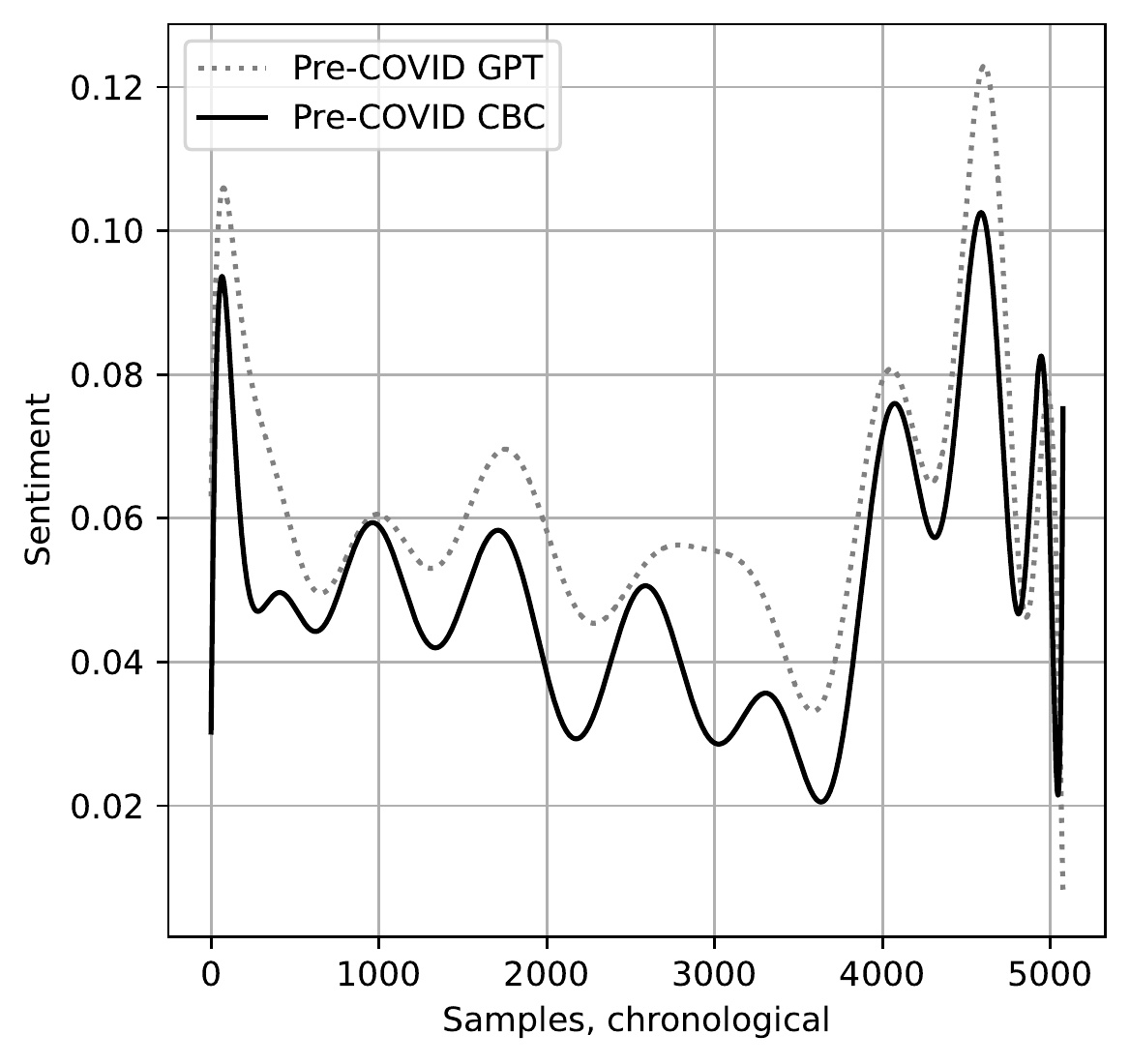}
  \caption{Correlation of sentiment in pre-COVID CBC and GPT articles over a ten year period. }
  \label{figure:control}
\end{figure}

\paragraph{Measure 4: Key Words}
\label{par:keywords}
We conduct a key word test on each respective corpus to identify repeatedly used terms bearing sentimental weight as per VADER. Following best practices, we only rank significant words in reference to each other rather than assigning significance to any term in isolation. We use the two formulae presented below for determining key words in a given corpus, as provided by \citet{rayson}.

We first calculate the averaged frequency $E_{i}$ for each word in our corpus with
\begin{equation}
E_{i}  = \frac{N_{i}\displaystyle\sum_{i}O_{i}}{\displaystyle\sum_{i}N_{i}}
\end{equation}
where $N$ is the total word count and $O$ is the frequency for the word.

Having now obtained a list of frequencies, we proceed with modifying our frequencies with a log-likelihood ($LL$) test:
\begin{equation}
LL = 2 \displaystyle\sum_{i}O_{i} \ln( \frac{O_{i}}{E_{i}} )
\end{equation}
We then rank our key words according to their respective LL values before comparing our two respective key word lists.

\section{Results}

\subsection{Fine-Tuning Validation}
Validating artificial text generation is a challenging task as there is no ``right'' answer when it comes to creating artificially generated text. Our primary goal in this case is to disambiguate whether our results are an effect of GPT-2 behavior (i.e. a result of model bias) or an effect of our fine-tuning and prompt engineering (i.e. a result of COVID-specific information). To do so, we first create a control dataset consisting of 5,077 randomly sampled CBC articles published prior to COVID between 14 January 2010 and 31 December 2019 (``pre-COVID CBC''). We then generate artificial articles with a standard context and a temperature of $0.5$. As shown below, our pre-COVID model produces articles whose distributions are highly statistically similar to the pre-COVID CBC data across our three primary measures, suggesting that any deviation from these levels of correlation in subsequent models is an effect of the COVID fine-tuning and not a default behavior of the model.

\paragraph{Sentiment}
We find fine-tuning GPT-2 with pre-COVID CBC data produces a model whose textual output is sentimentally similar to pre-COVID CBC articles as may be observed in \autoref{figure:control}. The sentiment distributions measured in our generated and real-world control datasets share an overlap of 97.7\% (Cohen's $d \approx 0.06$) and a moderately positive correlation coefficient of $r \approx 0.57$. We furthermore note GPT-2 is typically more positive in tone than CBC when examining the two distributions as a whole. A selection of validated sentences together with their respective sentiment values are presented in \autoref{table:sentences}.

\paragraph{Focus}
When measuring focus values for the control dataset, we again note a large overlap between GPT-2 and CBC at 93.6\% ($d \approx 0.16$) together with a weakly positive correlation or $r \approx 0.17$. These values suggest GPT-2 has learned focus trends latent in the pre-COVID CBC training set.

\paragraph{Key words}
Our final validation metric is a key words test using the process described in \autoref{par:keywords}. The mean log-likelihood of key words deployed by GPT-2 is $9.61$ ($95^{th}$ percentile $\approx 42$), indicating such terms are only marginally more likely to be used by GPT than by pre-COVID CBC.

\subsection{Measure 1: Sentiment}
\begin{figure}
  \centering
  \includegraphics[width=1\linewidth]{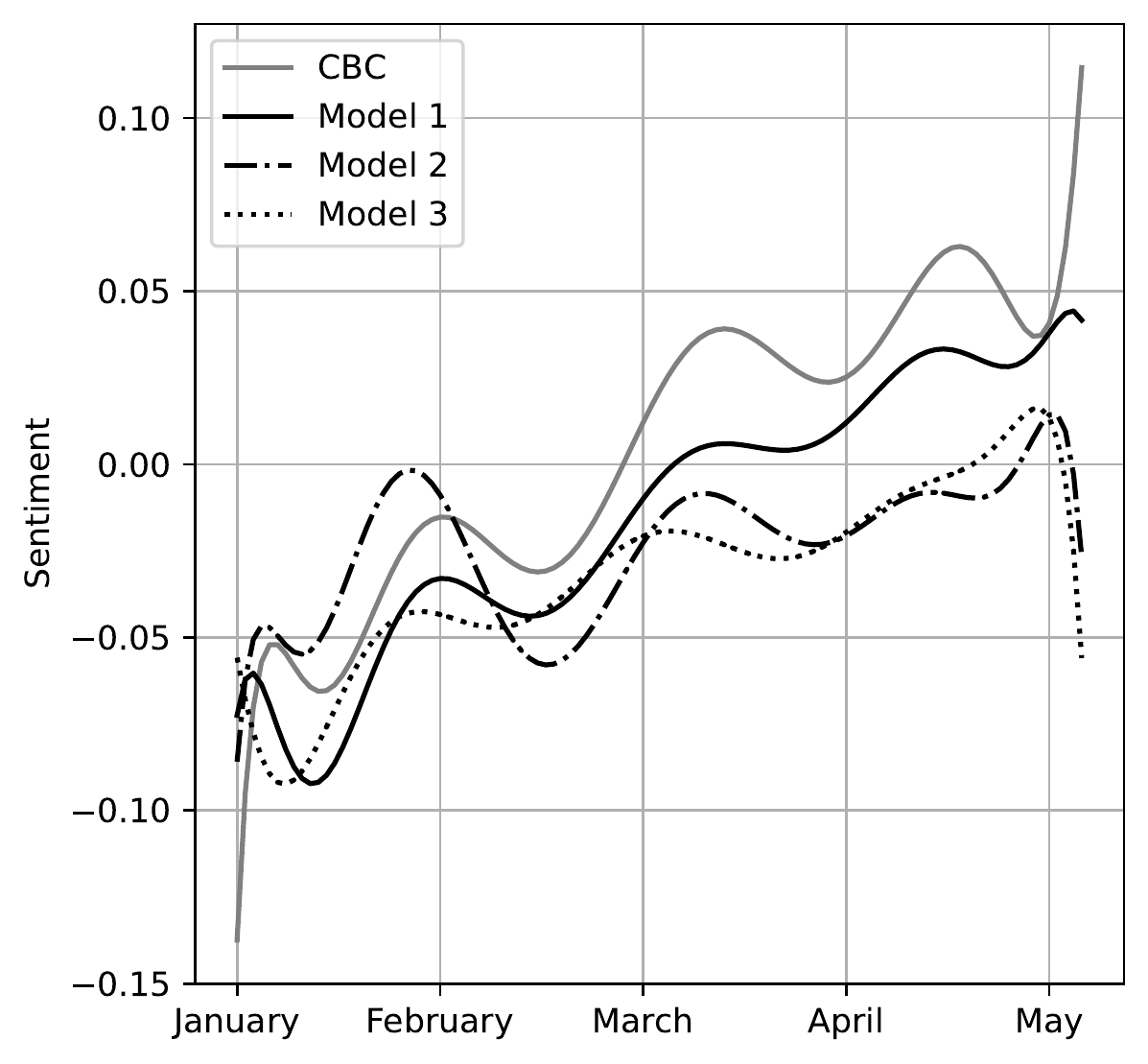}
  \caption{Averaged weekly article sentiment over the first four months of the pandemic.}
  \label{figure:sentiment}
\end{figure}
We begin by noting that CBC News' treatment of COVID-19 during our period of inquiry develops in two stages (\autoref{figure:sentiment}): articles prior to early March register overall as negative in their sentiment valence (stage 1), while articles written after the first two months become increasingly positive (stage 2). Note that in the pre-COVID data sentiment values were uniformly positive for both CBC and GPT.

When comparing our simulated texts to CBC, we find that our simulated corpora all demonstrate similar trends over time, but with significantly lower levels of positivity than the actual corpus, which is a direct reversal of the pre-COVID baseline. The headline-only model (Model 1) exhibits the highest level of correlation with the CBC corpus at $r \approx 0.28$, while the rolling context model (Model 3) exhibits the starkest overall difference in terms of generating more negative sentiment with an overall effect size of $d \approx -0.28$ (more than double what we see for Model 1 at $d \approx -0.12$). In general, we note that Model 1 adheres most strongly to CBC practices, while adding the CDC context, whether rolling or static, tends to make the models diverge more strongly from CBC practices.

\subsection{Measure 2: Named Entity Recognition}
\begin{figure}
  \centering
  \includegraphics[width=1\linewidth]{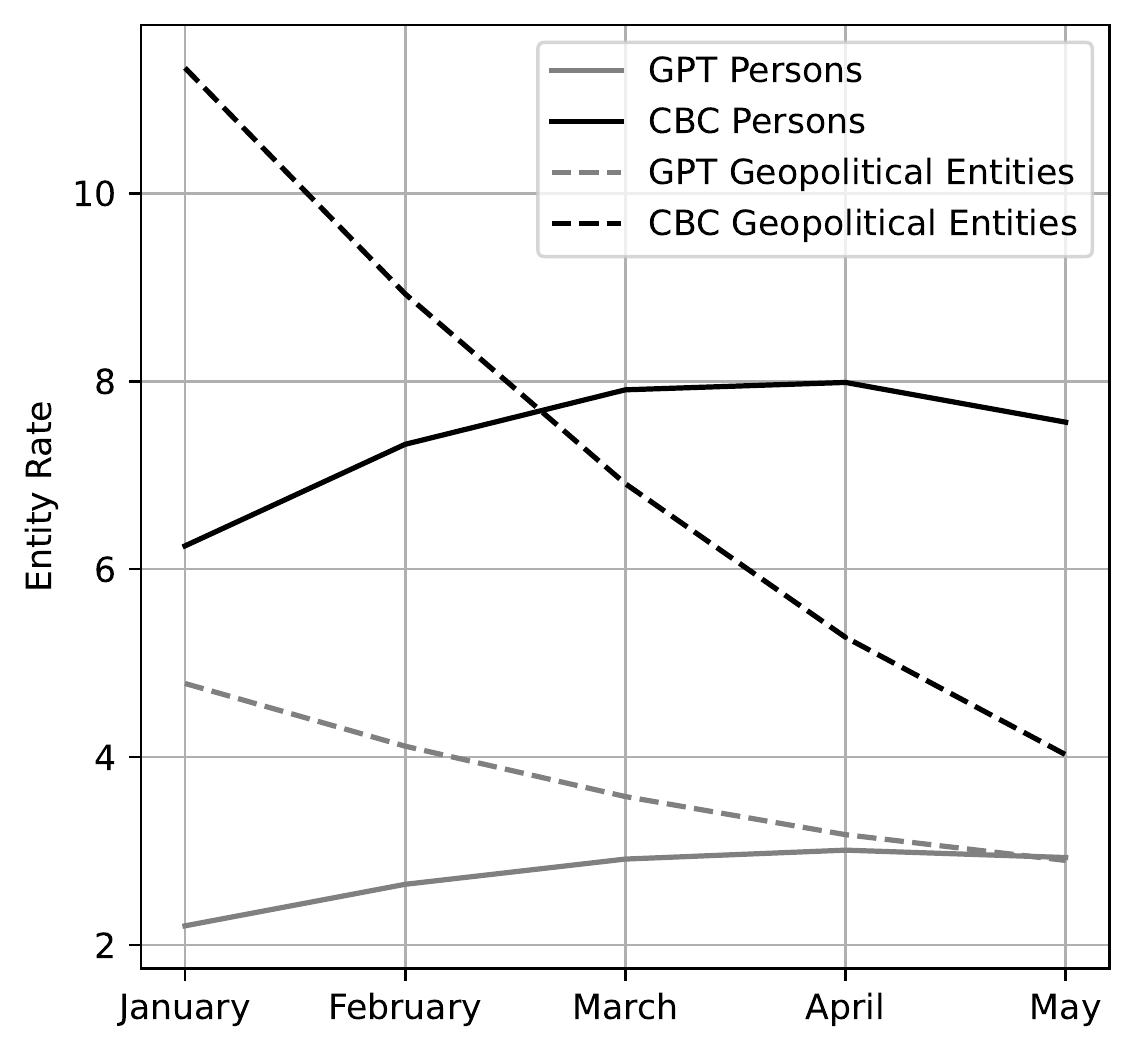}
  \caption{Average values of given entity types in CBC \& GPT articles over the first four months of the pandemic for Model 3.}
  \label{figure:ner}
\end{figure}
Tracking entities classified as persons, geopolitical entities (e.g., countries), and organizations (e.g., the World Health Organization), we find a similar two-stage process as we did when observing sentiment. As is observable in \autoref{figure:ner}, we see a significant decline of geopolitical entities after March in the CBC corpus replaced by a slight increase in notable persons. While the rolling context model exhibits decent correlation with the CBC corpus at $r \approx0.27$, we find a very strong discrepancy in the relative reliance on geopolitical entities in the GPT corpus compared to CBC with an overall effect size of $d \approx -0.63$.

\subsection{Measure 3: Focus}
Measuring the personal focus of articles reveals a number of trends. Predominant among these is a clear upwards trend over time in the CBC articles. Continuing the split stage analysis of the past measurements, we find focus increases linearly as the months of the pandemic pass. Higher focus values indicate that fewer entities are being discussed at greater length (i.e. are centralized more strongly). We also note a low correlation between article sentiment and article focus ($r \approx 0.18$), suggesting that focalization around fewer persons is associated with more positive messaging. We explore this effect further in \autoref{sec:centredness}.

In terms of our simulated corpus, we see that GPT-2 remains relatively consistent in both focus and sentiment over the course of our time window. While there remains an extremely weak positive correlation between sentiment and focus in the Model 3 corpora ($r \approx 0.02$), this is likely an artifact carrying over from the headlines themselves becoming more positive over time.

\subsection{Measure 4: Key Words}
When we observe the likelihood of a given word's appearance in one corpus or the other, here too we observe some notable trends.\footnote{We condition only on VADER vocabulary and not the full set of words.} We present a subsection of our results in \autoref{table:terms} using Model 3.

Conducting a qualitative analysis on the top key words underscores two points. First, we find Model 3 (and other models) routinely interpret COVID-19 as a flu, reflected in the model deploying terms like “flu” and “strain” more regularly than CBC News. This interpretation likely accounts for a majority of the discrepancies between the two corpora. Second, we find CBC is more likely to describe societal responses to COVID-19 (``emergency,'' ``crisis''), whereas GPT-2 draws on imagery to convey the medical threat of the disease (``sickened,'' ``infected'').
{\renewcommand{\arraystretch}{1.1}
\begin{table}
\centering
\begin{tabular}{|ll|ll|}
\hline
\textbf{CBC News} & LL & \textbf{GPT-2} & LL\\
\hline
“crisis” & 475 & “flu” & 2465 \\
“care” & 431 & “strain” & 871 \\
“cancelled” & 371 & “infected” & 855 \\
“isolation” & 363 & “great” & 558 \\
“emergency” & 264 & “sickened” & 400 \\
“anxiety” & 170 & “threat” & 317 \\
“support” & 158 & “cancer” & 302 \\
“sick” & 149 & “natural” & 249 \\
“critical” & 138 & “killed” & 189 \\
“vulnerable” & 134 & “dangerous” & 167 \\
\hline
\end{tabular}
\caption{A selection of the ten most prevalent sentimentally-charged terms in either corpus.}
\label{table:terms}
\end{table}}

\section{Discussion}
In this section we identify three noteworthy discrepancies between the behavior of our models and the real-world CBC corpus and discuss their potential implications.

\subsection{Effect 1: Positivity Bias (``Rally-Around-The-Flag'')}
We note that all of our simulated models trained on the COVID data generated news that was far more negative than actual coverage, which grew increasingly positive over time. This result is especially notable given that pre-COVID models were uniformly more positive than actual CBC articles.

A relevant theory that can help make sense of this is the “rally-around-the-flag” effect, which posits that national discourse trends in favour of reigning governments during times of crisis \citep{van2021covid}. Theorists in communication studies note news media do not remain neutral during crises, but instead work to assuage public fears by promoting trust in local leaders \citep{quandt2021stooges}.

The “rally-around-the-flag” effect could help explain why CBC News articles became more positive as lockdowns began and why our language models, which were not subject to such pressures, nevertheless remained more negative.  Regardless of the cause of this discrepancy between our models and CBC, it is worth noting our language models consistently interpreted COVID in more negative terms than this particular public broadcaster. An important aspect to underscore is that we do not see the same effect when we run the same process on a random assortment of pre-COVID articles, meaning our GPT models are not intrinsically more negative but rather interpret these particular events more negatively than CBC.

\subsection{Effect 2: Early Geopolitical Bias}
As we saw in \autoref{figure:ner}, CBC News relied on considerably more geopolitical entities in the early weeks of the pandemic than in the latter weeks, an effect our models only mildly reproduced. The strong decline of geopolitical entities in the CBC data past February suggests an editorial re-orientation away from understanding the pandemic in global geopolitical terms and towards a local health emergency that is more in line with what our models were producing from the beginning.

\subsection{Effect 3: Person versus Disease Centredness}
\label{sec:centredness}
\begin{figure}
  \centering
  \includegraphics[width=1\linewidth]{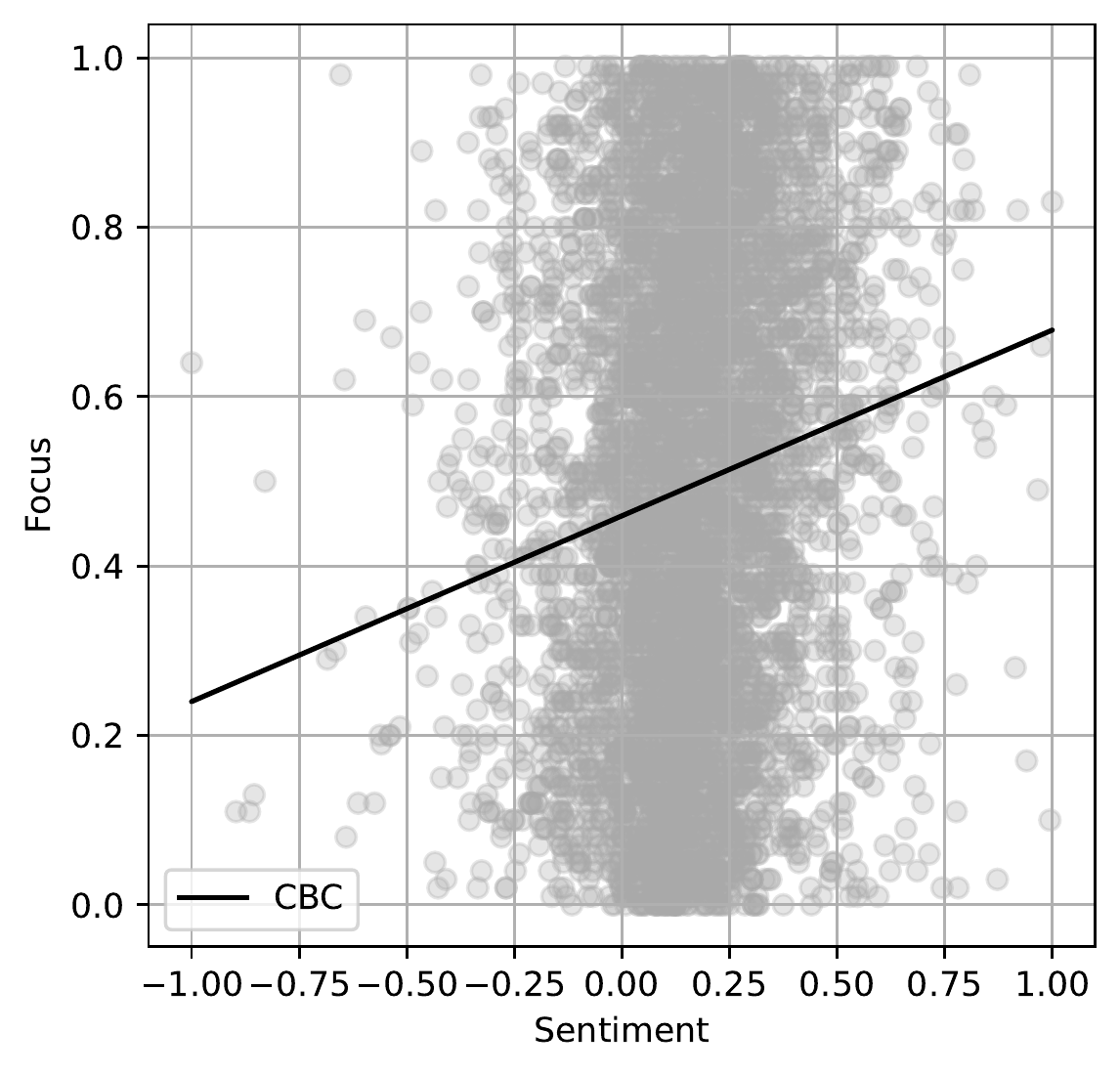}
  \caption{Relationship between focus and sentiment in CBC articles. Focus values are normalized.}
  \label{figure:focus_versus_sentiment}
\end{figure}
While the rate of geopolitical entities in the CBC data eventually converges with our GPT models, we see that the reliance on individual persons is consistently stronger in the CBC , something we did not see in the pre-COVID models. In conjunction with our key-word findings, this suggests that GPT's interpretation of the pandemic is far more medical and health-oriented (``infected,'' ``sickened'') than CBC, whose treatment remained more focused on people. As we show in \autoref{figure:focus_versus_sentiment}, this person-centredness is also associated with higher levels of positivity. Future work will want to explore whether this personal focalization was unique to the CBC, COVID, or the experience of social upheaval more generally.

\section{Conclusion}
The aim of our paper has been to develop a framework for using the text-generation affordances of large language models to better understand the interpretive perspectives of the news media when covering major social events. We rely on a simulative process whereby the generation of thousands of alternative views of a real world event can provide a framework for understanding the interpretive perspectives employed by news organizations.

Given that language models can approximate human discourse \citep{radford_language_nodate}, they can be used to generate a distribution of possible responses to an event to better understand the actual selection mechanisms used by real-world actors. Key to this process is validating the extent to which the qualities of artificially generated text are a function of model parameters or the process of fine-tuning, i.e. an effect of the real-world event we aim to simulate. Our aim in doing so is to illustrate how language models can be used as diagnostic tools for human behavior. Given no prior knowledge of a major event, what would a language model say? And what might this tell us about our own human reactions?

Based on the results we have obtained here, we see the following possible avenues for further research using LLMs for textual simulation:

\paragraph{Further Domain Exploration.} What other scenarios might LLMs be analytically useful for? In this paper, we have explored LLMs as a tool to assess media coverage, but future work will want to observe how they behave in other domains. News is a particularly well-structured form of textual communication and thus we expect LLMs to perform more adequately in this domain given prior research \citep{ueda_structured_2021}. We await future work exploring other textual domains.

\paragraph{Modeling Audience Expectations.} We have used GPT as a tool to assess the interpretive frameworks of the news media, specifically the CBC. However, we might also consider the ways in which LLMs can provide us with population-level expectations about an event. For example, the strong reliance on the ``flu'' in our models could be seen as a faithful mirror of how laypeople generally have thought about COVID (in distinction from public health experts). While one might argue that this is ``erroneous'' from a public health perspective, such semantic frameworks may be useful resources in fashioning public communication during times of crisis or upheaval. LLMs may be able to help us better understand what biases audiences are bringing to novel events thus helping experts craft more appropriate messaging that aligns with audience expectations.

\paragraph{Predicting Future Outcomes.} While we have used GPT as a tool to assess past behavior, future work could explore the predictive power of LLMs, while exercising a great deal of caution when it comes to their application. For example: Can LLMs identify future valuable research questions? Financial or economic events given changing real-world information? Or potential political crises given the communicative behavior of principal actors (e.g. politicians)? An equally potent line of research will want to explore the dangers of such approaches as in past experiences of predictive policing. New technologies always bring an admixture of analytical affordance and risk that needs to be better understand with respect to LLMs. More experimentation with respect to the efficacy of textual simulation is definitely warranted.

\bibliography{custom}
\bibliographystyle{acl_natbib}

\appendix
\section{Article Samples}
\label{sec:appendix}

We present further samples of both CBC and GPT-written articles concerning COVID-19:

\paragraph{Headline 3:} "Air Canada suspends all flights to Beijing and Shanghai amid coronavirus outbreak."

\subparagraph{CBC:} "Airlines around the world have started to cancel more flights to China as coronavirus fears have dragged down demand for air travel in the area. Air Canada normally runs 33 flights a week to China out of Toronto, Vancouver and Montreal but on Wednesday the airline announced it would halt all flights to Beijing and Shanghai until the end of February at least. "Air Canada's last flights departing Canada will operate today and the return flights will operate from Beijing and Shanghai tomorrow, January 30, 2020," the airline said. "Affected customers will be notified and offered options, including travel on other carriers where available, or a full refund. Air Canada regrets this situation and apologizes for the serious disruption to our customers' travel plans," the airline said in a statement. It isn't the only airline scaling back. British Airways on Wednesday suspended all direct flights to and from mainland China through to the end of February, although it will maintain limited service to Hong Kong..."

\subparagraph{GPT:} "Toronto's processing plant for large appliances grounded all Air Canada flights heading to and from Shanghai after Shanghai Municipal Airport in China confirmed that there has been a record number of cases of coronavirus there this year. The grounding applies to all international flights  including incoming and outgoing Air Canada and Air Canada Plus flights  to and from Shanghai. China is seeing a soaring number of avian coronavirus infections  at one point making it the top of the list of infectious diseases to watch out for. There is a recent trend of high-pathients experiencing shedding of viral loads, Peter Clark, assistant regional director of the Shanghai Centers for Disease Control and Prevention, told CBC News. This could be a result of sharing among regions of the same influenza virus circulation area...'"

\paragraph{Headline 4:} "Officials in northern Sask. link COVID-19 cases with travel to Alberta."

\subparagraph{CBC:} "Provincial and Indigenous health officials in northern Saskatchewan are investigating COVID-19 cases they say are linked to travel to Alberta. A statement from the Saskatchewan Health Authority says it and the Northern Inter-Tribal Health Authority have begun a contact tracing investigation into new cases of the novel coronavirus in the province's north that are linked to cross-boundary travel. The statement notes there have been reports of a number of COVID-19 cases at an oil sands facility north of Fort McMurray, Alta., although it doesn't specify whether any of the new cases in Saskatchewan are believed to be connected to that outbreak...."

\subparagraph{GPT:} "Northern Saskatchewan officials are linking travel to Alberta to clusters of a cluster infection. Health officials in Edmonton and Calgary are also investigating cases that are linked to travel to the Alberta province. Officials in northern Sask. are linking travel to Alberta to clusters of COVID-19, a coronavirus that's ravaged South Asia in the last 30 years. Health officials in Edmonton and Calgary are also investigating cases that are linked to travel to the Alberta province. (CBC) Minister of Health Emily Lackord says all of Canada is susceptible, but the outbreak is most aggressive yet. Saskatchewan's northern boundary is the only place with the disease, she says...'"

\end{document}